\documentclass[sigconf]{acmart}
\AtBeginDocument{%
  }

\copyrightyear{2025}
\acmYear{2025}
\setcopyright{acmlicensed}\acmConference[MM '25]{Proceedings of the 33rd ACM International Conference on Multimedia}{October 27--31, 2025}{Dublin, Ireland}
\acmBooktitle{Proceedings of the 33rd ACM International Conference on Multimedia (MM '25), October 27--31, 2025, Dublin, Ireland}
\acmDOI{10.1145/3746027.3755175}
\acmISBN{979-8-4007-2035-2/2025/10}
\usepackage{multirow}

\usepackage{xcolor}  % 提供颜色支持
\usepackage{colortbl}  % 提供表格颜色支持
\definecolor{myColor}{RGB}{247, 224, 213}  % 自定义粉色

\usepackage{hyperref}

\begin{document}

\title{Stepwise Decomposition and Dual-stream Focus: A Novel Approach for Training-free Camouflaged Object Segmentation}

\author{Chao Yin}
\email{yincaho@shu.edu.cn}
\orcid{0009-0005-9565-1571}
\affiliation{%
  \institution{Shanghai University}
  \city{Shanghai}
  \country{China}
}

\author{Hao Li}
\email{lihao2022@iscas.ac.cn}
\orcid{0000-0002-8827-8351}
\affiliation{%
  \institution{University of the Chinese Academy of Sciences}
  % \institution{Institute of Software, Chinese Academy of Sciences}
  \city{Beijing}
  \country{China}}

\author{Kequan Yang}
\email{kqyang@shu.edu.cn}
\orcid{https://orcid.org/0000-0002-5084-3474}
\affiliation{%
  \institution{Shanghai University}
  \city{Shanghai}
  \country{China}
}

\author{Jide Li}
\email{iavtvai@shu.edu.cn}
\orcid{0000-0002-0754-5842}
\affiliation{%
  \institution{Shanghai University}
  \city{Shanghai}
  \country{China}
}

\author{Pinpin Zhu}
\email{zhupp@shu.edu.cn}
\orcid{0009-0005-9781-5429}
\affiliation{%
  \institution{Shanghai University}
  \city{Shanghai}
  \country{China}
}

\author{Xiaoqiang Li}
\authornote{Corresponding author}
\email{xqli@shu.edu.cn}
\orcid{https://orcid.org/0000-0001-7243-2783}
\affiliation{%
  \institution{Shanghai University}
  \city{Shanghai}
  \country{China}
}

\renewcommand{\shortauthors}{Chao Yin et al.}

\begin{abstract}
  While promptable segmentation (\textit{e.g.}, SAM) has shown promise for various segmentation tasks, it still requires manual visual prompts for each object to be segmented. In contrast, task-generic promptable segmentation aims to reduce the need for such detailed prompts by employing only a task-generic prompt to guide segmentation across all test samples. However, when applied to Camouflaged Object Segmentation (COS), current methods still face two critical issues: 1) \textit{\textbf{semantic ambiguity in getting instance-specific text prompts}}, which arises from insufficient discriminative cues in holistic captions, leading to foreground-background confusion; 2) \textit{\textbf{semantic discrepancy combined with spatial separation in getting instance-specific visual prompts}}, which results from global background sampling far from object boundaries with low feature correlation, causing SAM to segment irrelevant regions. To mitigate the issues above, we propose \textbf{RDVP-MSD}, a novel training-free test-time adaptation framework that synergizes \textbf{R}egion-constrained \textbf{D}ual-stream \textbf{V}isual \textbf{P}rompting (RDVP) via \textbf{M}ultimodal \textbf{S}tepwise \textbf{D}ecomposition Chain of Thought (MSD-CoT). MSD-CoT progressively disentangles image captions to eliminate semantic ambiguity, while RDVP injects spatial constraints into visual prompting and independently samples visual prompts for foreground and background points, effectively mitigating semantic discrepancy and spatial separation. Without requiring any training or supervision, RDVP-MSD achieves a state-of-the-art segmentation result on multiple COS benchmarks. The codes will be available at \href{https://github.com/ycyinchao/RDVP-MSD}{https://github.com/ycyinchao/RDVP-MSD}.
\end{abstract}

\begin{CCSXML}
<ccs2012>
   <concept>
       <concept_id>10010147</concept_id>
       <concept_desc>Computing methodologies</concept_desc>
       <concept_significance>500</concept_significance>
       </concept>
   <concept>
       <concept_id>10010147.10010178.10010224.10010245.10010247</concept_id>
       <concept_desc>Computing methodologies~Image segmentation</concept_desc>
       <concept_significance>500</concept_significance>
       </concept>
   <concept>
       <concept_id>10010147.10010178.10010224.10010225.10010227</concept_id>
       <concept_desc>Computing methodologies~Scene understanding</concept_desc>
       <concept_significance>500</concept_significance>
       </concept>
 </ccs2012>
\end{CCSXML}

\ccsdesc[500]{Computing methodologies}
\ccsdesc[500]{Computing methodologies~Image segmentation}
\ccsdesc[500]{Computing methodologies~Scene understanding}

\keywords{Training-free Camouflaged Object Segmentation, Promptable Segmentation, Binary Segmentation}


\maketitle

\section{Introduction}
\label{sec: introduction}

\begin{figure}[t]
\centering
\includegraphics[width=0.8\columnwidth]{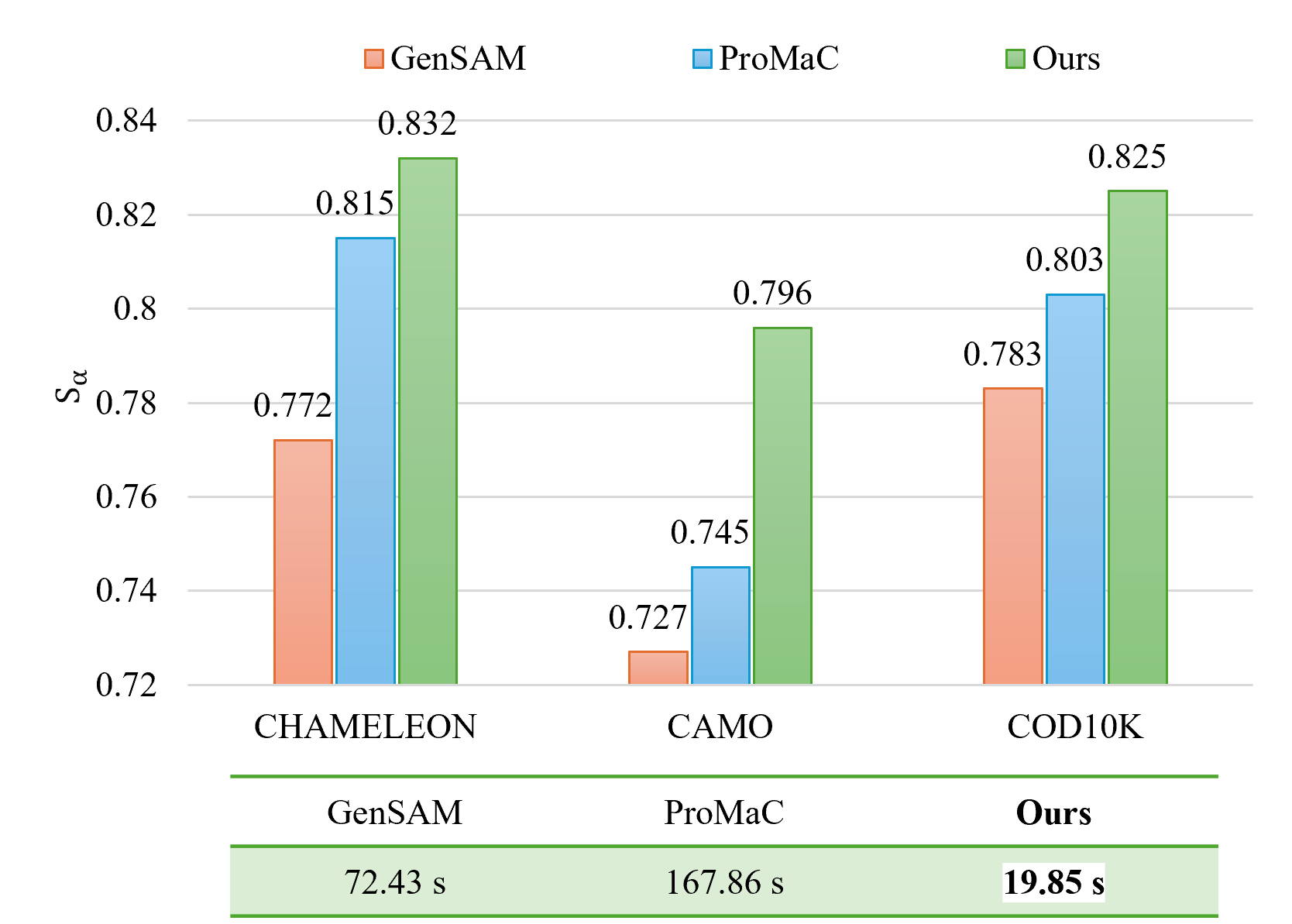}
\caption{\textbf{Superior Performance and Efficiency:} The proposed RDVP-MSD achieves state-of-the-art performance while requiring only about $19.85s$ per image, outperforming existing approaches (GenSAM \cite{hu2024relax}/ProMaC \cite{hu2024leveraging}) by $+7.6\%/3.7\%~S_\alpha$ and $3.6\times/8.5\times$ speedup (averaged across all datasets). Results on three benchmark datasets demonstrate consistent advantages in both accuracy and computational efficiency.}
\label{fig: qualitative}
\end{figure}

\begin{figure*}[t]
\centering
\includegraphics[width=0.9\textwidth]{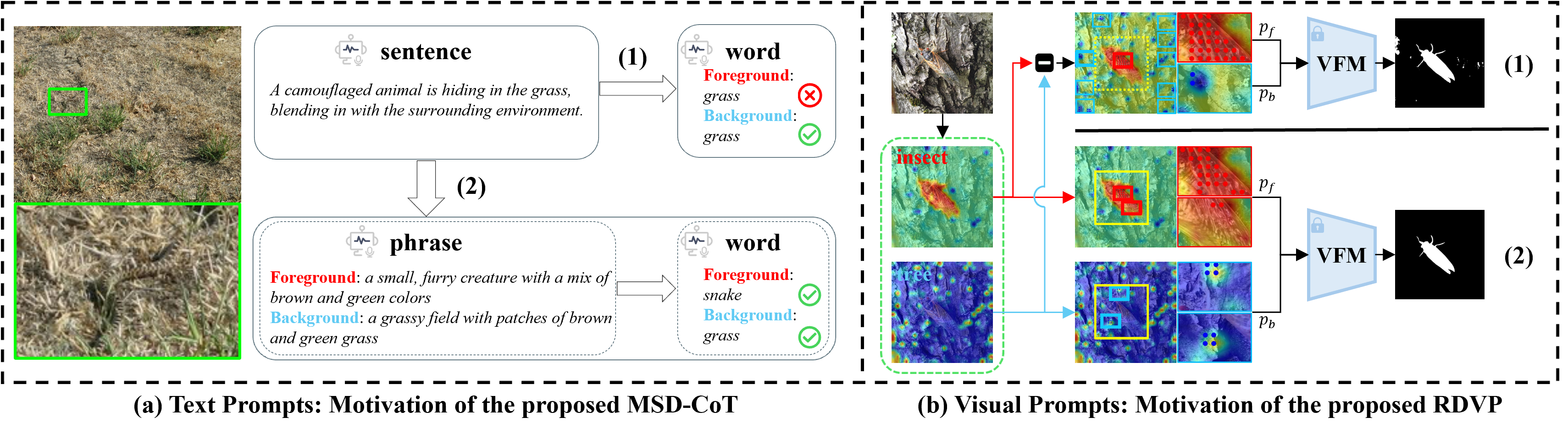} % Reduce the figure size so that it is slightly narrower than the column.
\caption{
\textbf{Motivation of the proposed RDVP-MSD.} 
\textbf{(a) Text Prompts}: 
(1) Existing methods directly extract category cues from image captions (\textit{e.g.}, ``A camouflaged animal is hiding in the grass...''), yielding erroneous instance-specific text prompts (\textit{e.g.}, foreground: ``grass'' $\rightarrow$ misclassified).  
(2) Our MSD-CoT introduces a phrase-level disentanglement stage to decouple entangled semantics, purifying foreground (``snake'') and background (``grass'') prompts via MLLM-guided semantic disambiguation;  
\textbf{(b) Visual Prompts}: 
(1) Prior approaches depend on consistency heatmaps (foreground $-$ background), where globally sampled background points (blue) introduce semantic discrepancy and spatial separation, misguiding models to segment irrelevant regions.  
(2) The proposed RDVP independently selects high-confidence foreground (red) / background (blue) points within the object bounding box, forcing VLM to focus on discriminative regions surrounding camouflaged objects.
}
\label{fig: motivation}
\end{figure*}
Camouflaged Object Segmentation (COS) confronts the critical challenge of precisely identifying and segmenting objects exhibiting high visual similarity with their surrounding environments. The inherent complexity of this task significantly amplifies annotation costs, with each pixel-level image-mask pair requiring approximately 60 minutes for manual annotation \cite{fan2020camouflaged}. While weak supervision paradigms \cite{he2023weakly,he2023weaklysam,niu2024minet,chen2025just} have been proposed to mitigate annotation intensity, their performance degrades progressively with increasing label sparsity. Recent advancements in Vision Foundation Models (VFMs), particularly those supporting promptable segmentation tasks (\textit{e.g.}, SAM \cite{kirillov2023segment}), demonstrate promising potential by achieving competitive segmentation accuracy through minimal manual instance-specific visual prompts (\textit{e.g.}, sparse point annotations). This breakthrough has catalyzed the emergence of automated promptable segmentation methodologies \cite{hu2024relax,hu2024leveraging,tang2024chain}, predominantly adopting a task-generic prompting strategy \cite{hu2024relax,hu2024leveraging} where a single task-generic prompt (\textit{e.g.}, "\textit{camouflaged object}") is indiscriminately applied across all test samples within a target domain (\textit{e.g.}, \textit{COS}).

Existing approaches for generating VFM-compatible instance-specific visual prompts, exemplified by GenSAM \cite{hu2024relax}, employ a cyclic-generation mechanism that iteratively extracts instance-specific visual prompts through Multimodal Large Language Models (MLLMs) \cite{li2023blip,openai2024gpt4v,liu2023visual,liu2024improved}, coupled with Vision-Language Models (VLMs, \textit{e.g.}, Spatial CLIP \cite{hu2024relax}). However, these methods exhibit fundamental limitations in complex scene understanding, particularly when target objects exhibit complete visual similarity with background textures. ProMaC \cite{hu2024leveraging} alleviates this challenge by strategically leveraging hallucination priors, yet its reliance on multi-patch visual question-answering via MLLMs to filter irrelevant hallucinations introduces substantial computational overhead. As shown in Figure \ref{fig: qualitative}, although ProMaC achieves higher performance than GenSAM, this comes at the cost of sacrificing the efficiency of single-image inference. Our proposed method not only achieves better accuracy but also significantly improves the efficiency of single-image inference (outperforms GenSAM and ProMaC by+7.6\% and+3.7\% on the $S_{a}$ metric while being 3.6$\times$/8.5$\times$ faster).

Contemporary task-generic promptable segmentation methods \cite{hu2024relax,hu2024leveraging} confront two principal problems in camouflaged scene understanding. First, existing methods that directly derive instance-specific text prompts from holistic image captions suffer from unresolved semantic ambiguity. As illustrated in Figure \ref{fig: motivation}(a), camouflaged image captions (\textit{e.g.}, "\textit{A camouflaged animal is hiding in the grass, blending in with the surrounding environment.}") contain insufficient discriminative cues, frequently inducing foreground-background confusion (\textit{i.e.}, misidentifying background elements (\textit{e.g.}, "\textit{grass}") as a foreground text prompt). Our framework introduces a phase of phrase disentanglement that models contextual coexistence patterns between camouflaged objects and backgrounds. Through linguistic stepwise construction, this mechanism decomposes a holistic caption into a foreground phrase ("\textit{a small, furry creature with a mix of brown and green colors}") and a background phrase ("\textit{a grassy field with patches of brown and green grass}"), followed by semantic purification via progressive MLLM interrogation to distill noise-free keywords (\textit{e.g.}, foreground: "\textit{snake}"). Second, previous visual prompting strategies \cite{hu2024relax,hu2024leveraging} based on the consistency heatmap (foreground $-$ background) suffer from semantic discrepancy and spatial separation, as illustrated in Figure \ref{fig: motivation}(b). Specifically, global background sampling introduces semantic discrepancy (low feature correlation with the camouflaged object) and spatial separation (sampling points distant from object boundaries), causing VFMs to segment spurious regions. Our framework innovatively generates independent foreground/background heatmaps within the object bounding box, strategically selecting foreground points through corresponding instance-specific text prompts alignment maximization while sampling adversarial background points that exhibit high semantic response to background cues and spatial adjacency for focusing on the interior of the camouflaged object.

In light of the issues above, we propose RDVP-MSD, a novel training-free test-time adaptation framework that synergizes \textbf{R}egion-constrained \textbf{D}ual-stream \textbf{V}isual \textbf{P}rompting (RDVP) via \textbf{M}ultimodal \textbf{S}tepwise \textbf{D}ecomposition Chain of Thought (MSD-CoT) as shown in Figure \ref{fig: framework}. The MSD-CoT mechanism implements a four-step stepwise reasoning process: 1) Caption Generation, 2) Phrase Disentanglement, 3) Keyword Identification, and 4) Coarse Location. This structured decomposition effectively mitigates semantic ambiguities inherent in task-generic promptable methods, reducing the misclassification. Complementing the linguistic refinement, the Text-to-Mask Generator employs RDVP, which injects spatial constraints into visual prompting and independently samples high-confidence foreground/background points within bounding boxes. In the coarse stage, the Text-to-Mask Generator leverages phrase-level text prompts to generate coarse instance-level visual prompts via RDVP, which are fed into the VFM to produce initial segmentation masks. These masks are then refined into tighter bounding boxes. The fine-grained stage inherits word-level text prompts and the refined boxes, applying the same RDVP processing to focus on microscopic texture contrasts for pixel-accurate segmentation.

As illustrated in Table \ref{tab:sota}, leveraging MSD-CoT and RDVP, our proposed RDVP-MSD outperforms the state-of-the-art weakly supervised methods (point/scribble annotations) in Camouflaged Object Segmentation (COS). Notably, RDVP-MSD surpasses all existing task-generic promptable methods in COS benchmarks while maintaining a zero-training regime. Our principal contributions are threefold: (1) The proposed training-free test-time adaptation framework RDVP-MSD enables precise camouflaged object segmentation with faster inference speeds (8.5$\times$ faster than ProMaC); (2) Multimodal Stepwise Decomposition Chain of Thought (MSD-CoT), which mitigates semantic ambiguity in task-generic promptable methods through progressive sentence-phrase-word decomposition; (3) Region-constrained Dual-stream Visual Prompting (RDVP), which independently acquires adaptive foreground/background points within object bounding boxes, forcing VFM to focus on microscopic texture contrasts around camouflaged objects.

\section{Related Work}
\subsection{Camouflaged Object Segmentation}
Camouflaged Object Segmentation (COS) \cite{fan2020camouflaged,yin2024camouflaged,yin2025dual} is the task of identifying and segmenting objects that exhibit high visual similarity to their background, making them difficult to identify and segment. This task is highly challenging due to the complex nature of camouflage. It has substantial practical applications in areas such as military surveillance \cite{truong2024style,wang2024edge,liu2025multi}, wildlife monitoring \cite{zhai2024green,wang2024depth,liu2025improving}, and autonomous driving \cite{zhu2025camoenv}. Accurately segmenting camouflaged objects is essential for systems requiring high precision in object recognition and scene understanding.

Until now, various methods have been proposed to enhance COS performance, often relying on auxiliary information such as edge features \cite{he2023camouflaged,sun2022bgnet,zhao2024focusdiffuser}, frequency domain \cite{wang2025multidimensional,sun2024frequency,xie2023frequency}, depth \cite{liu2024depth,wu2023source,wu2023object}, and gradient cues \cite{ji2023deep}. These approaches have been instrumental in advancing COS but are predominantly designed within a supervised learning framework. The reliance on pixel-level annotations makes these methods highly resource-intensive and not scalable. Additionally, incorporating auxiliary information often requires extra annotations or depends on pre-trained models, limiting their adaptability to new, unseen data. To alleviate the annotation burden, researchers have explored semi-supervised \cite{fu2024semi,zhang2024learning,lai2024camoteacher} and weakly supervised \cite{niu2024minet,chen2025just,chen2024sam} (\textit{e.g.}, scribbles, points, or bounding boxes) learning methods. While these approaches reduce the need for extensive manual annotation, they typically encounter performance trade-offs as the level of supervision weakens. As supervision shifts from complete annotation to weaker signals, the segmentation accuracy can degrade, especially when methods are transferred across domains or tasks. Despite these challenges, the continued refinement of these methods strives to balance reduced supervision with high segmentation performance, advancing the potential of COS to be performed with minimal manual intervention.

\subsection{Segment Anything Model for COS}
The application of the Segment Anything Model (SAM \cite{kirillov2023segment}) in the domain of COS has been a significant development, driven by the increasing ability of models to perform segmentation with minimal supervision. SAM \cite{kirillov2023segment}, initially designed for more generic segmentation tasks, has shown promise when extended to COS. However, researchers \cite{ji2023sam,zhou2024multi,ji2024correction} have found that directly generalizing SAM to COS often leads to unsatisfactory results, as camouflaged objects exhibit high visual similarity with their backgrounds. Early attempts \cite{zhou2024multi,chen2023sam,zhang2025comprompter} to adapt SAM for COS involved using fully supervised masks for fine-tuning an adapter model to improve performance in camouflaged scenarios. Other approaches have explored using weaker supervision signals \cite{he2023weaklysam,chen2024sam,yu2024exploring}, such as pseudo-labeling through SAM, to generate training data for further model refinement. Despite the advancements, these methods still rely on manual instance-specific visual prompts, which require extensive manual effort. Recently, a shift toward train-free test-time adaptation methods \cite{hu2024relax,hu2024leveraging} has emerged, offering a significant breakthrough. These approaches enable the model to automatically generate instance-specific visual prompts from a task-generic prompt without fine-tuning or supervision.

\begin{figure*}[t]
\centering
\includegraphics[width=1.0\textwidth]{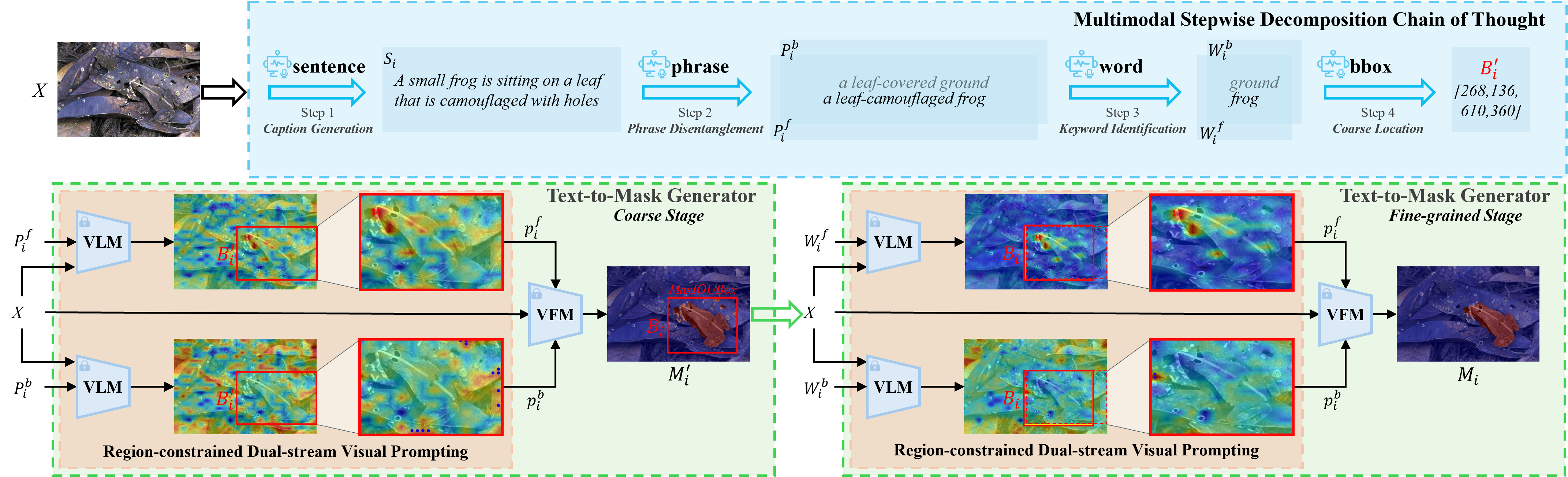} % Reduce the figure size so that it is slightly narrower than the column.
\caption{
Overview of the proposed RDVP-MSD framework. It comprises two core components: (1) Multimodal Stepwise Decomposition Chain of Thought (MSD-CoT), which applies a sentence-phrase-word decomposition strategy in a four-step stepwise reasoning process: Caption Generation, Phrase Disentanglement, Keyword Identification, and Coarse Location. This process progressively refines image captions into disentangled instance-specific text prompts, mitigating foreground-background ambiguity and associating bounding boxes with the corresponding semantic regions. (2) Text-to-Mask Generator, which employs Region-constrained Dual-stream Visual Prompting (RDVP) in a coarse-to-fine manner. Initially, coarse masks are generated using phrase-level text prompts, followed by refinement using discriminative word-level text prompts to achieve high-precision segmentation.
}
\label{fig: framework}
\end{figure*}

\section{Method}
\subsection{Framework Overview}
As illustrated in Figure. \ref{fig: framework}, our proposed RDVP-MSD is a training-free test-time adaptation framework for segmenting camouflaged objects with only a single task-generic prompt. Specifically, given an image $X\in \mathbb{R}^{H \times W \times 3}$ containing the camouflaged scene from a test set, the RDVP-MSD generates a corresponding segmentation mask $M \in \mathbb{R}^{H \times W}$ under the task-generic prompt $P_g$ (\textit{e.g.}, \textit{camouflaged object}, \textit{camouflaged animal}, \textit{camouflaged entity}.) setting by synergizing three frozen pre-trained models: a Multimodal Large Language Models (MLLMs, \textit{e.g.}, LLaVA \cite{liu2023visual,liu2024improved}) for instance-specific text prompts generation, a Vision-Language Models (VLMs, \textit{e.g.}, Spatial CLIP \cite{hu2024relax}) for text-to-visual prompt conversion, and a promptable Visual Foundation Models (VFMs, \textit{e.g.}, SAM \cite{kirillov2023segment}) for mask prediction. This process eliminates manual prompts and fine-tuning while maintaining generalization across diverse camouflaged scenarios.

In RDVP-MSD, only having MLLM, VLM, and VFM, despite their powerful capabilities, may still be insufficient to handle the COS task effectively. Therefore, we propose the Multimodal Stepwise Decomposition Chain of Thought (MSD-CoT) that progressively decomposes into hierarchical instance-specific text prompts: (1) phrase-level text prompts (\textit{e.g.}, "\textit{a leaf-camouflaged frog}" and "\textit{a leaf-covered ground}") to model foreground-background disentanglement and (2) word-level text prompts (\textit{e.g.}, "\textit{frog}" and "\textit{ground}"). To achieve refined segmentation, we introduce the Text-to-Mask Generator, which progressively refines the segmentation mask through hierarchical processing. Within this module, the proposed Region-constrained Dual-stream Visual Prompting (RDVP) employs the VLM to generate instance-specific visual prompts — foreground/background points are restricted to the predicted bounding box and are obtained separately from their respective heatmaps — which are then fed into the VFM for the mask generation. To mitigate uncertainty from stochastic MLLM outputs, we introduce the Self-Consistency Mask Selection, which generates multiple segmentation candidates in parallel under varying task-generic prompts and selects the most consistent mask as the final prediction via consensus voting.

\subsection{Text Prompt Generation}

The text prompt generation leverages the MLLM to transform a task-generic prompt $P_g$ into instance-specific text prompts tailored for each input image. Despite the advanced visual question capabilities of MLLMs, accurately generating instance-specific text prompts for camouflaged objects remains challenging due to their high visual similarity to surrounding backgrounds. Prior approaches \cite{hu2024relax} have attempted to mitigate these challenges by incorporating image captions as textual priors. Later studies \cite{hu2024leveraging} proposed using multiple local image patches to induce hallucinations in MLLMs for generating candidate knowledge, thereby reducing irrelevant hallucinations. However, as discussed in Section \ref{sec: introduction}, these methods typically incur substantial computational overhead. To overcome these limitations, we introduce a novel phrase disentanglement strategy integrated into the multimodal reasoning process, termed the Multimodal Stepwise Decomposition Chain of Thought (MSD-CoT). MSD-CoT explicitly models contextual coexistence between camouflaged objects and their backgrounds, effectively disentangling semantically entangled concepts and significantly enhancing the specificity and accuracy of the generated instance-specific text prompts.
\subsubsection{Multimodal Stepwise Decomposition Chain of Thought}
\label{sec: MSD-CoT}

Chain of Thought, a method leveraging intermediate reasoning steps generated by large language models (LLMs) to enhance task-solving capabilities, has demonstrated remarkable improvements in complex NLP reasoning tasks \cite{kojima2022large,wei2022chain}. Recently, this paradigm has been extended to multimodal large language models (MLLMs), achieving notable performance in visual-language understanding tasks \cite{zhang2024multimodal,mitra2024compositional,wang2024t}. However, studies \cite{zhang2024multimodal} have revealed that directly generating instance-specific text prompts from image captions often leads to significant information loss, which is particularly problematic in scenarios involving highly camouflaged objects. Hence, we argue that relying solely on image captions for deriving instance-specific text prompts is suboptimal and inadequate for precise recognition of highly camouflaged objects.

Inspired by the "\textit{Let's think step by step}" prompting strategy \cite{kojima2022large}, we propose a novel Multimodal Stepwise Decomposition Chain of Thought (MSD-CoT). MSD-CoT progressively obtains hierarchical instance-specific text prompts (phrase and word-level text prompts) through a structured sentence-phrase-word decomposition process, significantly improving multimodal reasoning accuracy. The MSD-CoT consists of four essential steps: Caption Generation, Phrase Disentanglement, Keyword Identification, and Coarse Location.
\paragraph{Caption Generation.} Initially, to strengthen the model’s understanding and querying capability regarding specific camouflaged objects, we employ an MLLM to generate a holistic scene description sentence $S_i$ from the input image:
\begin{align}
    S_i = MLLM (X, Q_i^s),
\end{align}
where $Q_i^s$ represents a task-specific query prompting the model, \textit{e.g.}, "\textit{This image is from $P_g^i$ detection task, describe the $P_g^i$ in one sentence.}" with $P_g^i \in P_g$. Here, $i$ denotes the number of repetitions, typically set to 3 by default (details in Section \ref{sec: maskSelection}).
\paragraph{Phrase Disentanglement.} Due to the inherent information loss in direct image-to-caption translations \cite{zhang2024multimodal}, directly generating word-level text prompts from captions often introduces semantic ambiguity, as exemplified in Figure \ref{fig: motivation}(a). To overcome this limitation, we introduce an intermediate phrase disentanglement stage. This stage explicitly models the contextual coexistence between the camouflaged object and its environment, disentangling the intertwined foreground-background semantics. It generates more discriminative phrase-level text prompts:
\begin{align}
    P_i^{f}, P_i^{b} = MLLM (X, Q^s_i, S_i, Q_i^p),
\end{align}
where $Q_i^p$ denotes a descriptive phrase query, instructing the model to "\textit{Provide a concise and comprehensive descriptive compound noun phrase for $P_g^i$ and its environment.}"
\paragraph{Keyword Identification.} The phrase-level disentanglement from the previous step explicitly forces the MLLM to differentiate key foreground and background features, enabling further semantic refinement. This step progressively decomposes the phrase-level text prompts into precise word-level representations:
\begin{align}
    W_i^{f}, W_i^{b} = MLLM (X, Q^s_i, S_i, Q^p_i, P_i^{f}, P_i^{b}, Q_i^w),
\end{align}
where $Q_i^w$ is a keyword identification query, such as "\textit{Name of the $P_g^i$ and its environment in one word.}"
\paragraph{Coarse Location.} Previous studies \cite{hu2024leveraging,liu2023visual,liu2024improved} have demonstrated that object categories generated by MLLMs can be associated with object regions through bounding box queries. We observed that for objects with lower camouflage levels, MLLMs typically generate relatively accurate bounding boxes that encompass the majority of the object. Conversely, for highly camouflaged objects, where visual features are difficult to distinguish from the background, we introduce a fault-tolerant mechanism. Specifically, we use an image-level bounding box as the initial coarse bounding box for providing a broader spatial constraint:
\begin{align}
    B_i^{'} = MLLM (X, Q^s_i, S_i, Q^p_i, P_i^{f}, P_i^{b}, Q_i^w, W_i^{f}, W_i^{b}, Q_i^{bbox}),
\end{align}
where $Q_i^{bbox}$ is an object bounding box query such as, "\textit{This image is from the $P_g^i$ detection task, output the bounding box of the $P_g^i$.}"

In summary, the proposed MSD-CoT effectively provides discriminative phrase-level and word-level text prompts along with preliminary bounding boxes. These hierarchical text prompts and coarse locations are subsequently utilized in the two Text-to-Mask Generators (as shown in Figure \ref{fig: framework}), significantly enhancing segmentation accuracy and robustness for the COS task.

\begin{table*} % * 表示这个表格在双栏模板中使用单栏
    \caption{Quantitative comparison across three standard benchmarks under different settings. $^{\times}$ denotes the absence of explicit mention and unavailable code in the corresponding paper. '↑' indicates higher is better, and '↓' indicates lower is better. The best results of different settings are highlighted in \textbf{bold}.}
  \centering

  \tabcolsep=0.013\linewidth

    \begin{tabular*}{\linewidth}{c|c|cccc|cccc|cccc}
    \toprule
         \multirow{2}{*}{Methods} & 
         \multirow{2}{*}{Venue} & 
         \multicolumn{4}{c|}{COD10K-TEST (2,026)} &
         \multicolumn{4}{c|}{CAMO-TEST (250)} & 
         \multicolumn{4}{c}{CHAMELEON (76)}

         \\ 
         \cline{3-14}
         &
         & $S_\alpha$↑ & $F_\beta$↑ & $M$↓ & $E_{m}^\phi$↑
         & $S_\alpha$↑ & $F_\beta$↑ & $M$↓ & $E_{m}^\phi$↑
         & $S_\alpha$↑ & $F_\beta$↑ & $M$↓ & $E_{m}^\phi$↑
        \\

    \midrule
    \multicolumn{13}{c}{\textbf{Point Supervision Setting}}\\
    \midrule
SS \cite{zhang2020weakly} & CVPR20                & .642   & .509   & .087   & .733   & .649   & .607   & .148   & .652      & .711   & .660   & .105   & .712\\
SCWS \cite{yu2021structure} & AAAI21              & .738   & .593   & .082   & .777   & .687   & .624   & .142   & .672      & .714   & .684   & .097   & .739\\
TEL \cite{liang2022tree}                       & CVPR22                                       & .724   & .633   & .057   & .826   & .717   & .681   & .104   & \textbf{.797}     & .785   & .708   & .073   & .827\\

CRNet \cite{he2023weakly} & AAAI23             & .711   & .607   & .060   & .802   & .663   & .629   & .137   & .688      & .725   & .688   & .092   & .746\\
SAM-P \cite{kirillov2023segment} & ICCV23             & .765   & .694   & .069   & .796   & .677   & .649   & .123   & .693      & .697   & .696   & .101   & .745\\
WS-SAM \cite{he2023weaklysam} & NeurIPS23         & \textbf{.790}   & \textbf{.698}   & \textbf{.039}   & \textbf{.856}   & \textbf{.718}   & \textbf{.703}   & \textbf{.102}   & .757      & \textbf{.805}   & \textbf{.767}   & \textbf{.056}   & \textbf{.868}\\

\midrule
    \multicolumn{13}{c}{\textbf{Scribble Supervision Setting}}\\
    \midrule

SS \cite{zhang2020weakly} & CVPR20                & .684   & .536   & .071   & .770   & .696   & .615   & .118   & .786      & .782   & .692   & .067   & .860\\
SCWS \cite{yu2021structure} & AAAI21              & .710   & .602   & .055   & .805   & .713   & .658   & .102   & .795      & .792   & .758   & .053   & .881\\
TEL \cite{liang2022tree}                       &      CVPR22                                        & .727   & .623   & .063   & .803   & .645   & .662   & .133   & .674       & .746   & .712   & .094   & .751\\
CRNet \cite{he2023weakly} & AAAI23            & .733   & .637   & .049   & .832   & .735   & .709   & .092   & .815       & .818   & .791   & .046   & .897\\
SAM-S \cite{kirillov2023segment} & ICCV23             & .772   & .695   & .046   & .828   & .731   & .682   & .105   & .774      & .650   & .729   & .076   & .820\\
WS-SAM \cite{he2023weaklysam} & NeurIPS23         & \textbf{.803}   & \textbf{.719}   & \textbf{.038}   & \textbf{.878}   & .759   & \textbf{.742}   & .092   & .818      & .824   & \textbf{.777}   & .046   & .897\\
WSMD \cite{zha2024weakly} & AAAI24              & .761   & .600   & .049   & .839   & \textbf{.793}   & .704   & \textbf{.079}   & \textbf{.866}      & .816   & .715   & .052   & .884\\
$^{\times}$MINet \cite{niu2024minet} & ACM MM24     & .749   & -      & .049   & .840   & .750   & -      & .091   & .840      & \textbf{.825}   & -      & \textbf{.044}   & \textbf{.910}\\
\midrule
    \multicolumn{13}{c}{\textbf{Task-Generic Prompt Setting}}\\
    \midrule

CLIP\_Surgey+SAM \cite{li2025closer} & PR25& .629   & .488   & .173   & .698   & .612   & .520   & .189   & .692      & .689   & .606   & .147   & .741\\
GPT4V+SAM \cite{openai2024gpt4v,kirillov2023segment} & Arxiv23        & .601   & .448   & .187   & .672   & .573   & .466   & .206   & .666      & .637   & .557   & .180   & .710\\
LLaVA1.5+SAM \cite{liu2024improved,kirillov2023segment} & CVPR24   & .662   & .530   & .170   & .728   & .501   & .401   & .314   & .585      & .666   & .561   & .168   & .718\\
X-Decoder \cite{zou2023generalized} & CVPR23         & .652   & .556   & .171   & .705   & .709   & .628   & .104   & .745      & .716   & .654   & .124   & .748\\
SEEM \cite{zou2024segment} & NeurlPS23           & .425   & .001   & .143   & .280   & .404   & .023   & .192   & .315      & .454   & .011   & .094   & .307\\
GroundingSAM \cite{ren2024grounded} & Arxiv24      & .764   & .670   & .085   & .813   & .707   & .656   & .157   & .753      & .744   & .662   & .122   & .776\\
GenSAM \cite{hu2024relax} & AAAI24            & .783   & .717   & .055   & .845   & .727   & .694   & .105   & .799      & .772   & .721   & .086   & .812\\
$^{\times}$MMCPF \cite{tang2024chain} & ACM MM24     & .733   & -      & .065   & -      & .749   & -      & .100   & -            & -      & -      & -      & -\\
ProMaC \cite{hu2024leveraging} & NeurIPS24         & .803   & .750   & .042   & .875   & .745   & .732   & .100   & .830      & .815   & .802   & .053   & .891\\
\rowcolor{myColor}  % 设置粉色背景
Ours                     & & \textbf{.825}   & \textbf{.775}   & \textbf{.038}   & \textbf{.877}   & \textbf{.796}   & \textbf{.785}   & \textbf{.081}   & \textbf{.848}      & \textbf{.832}   & \textbf{.814}   & \textbf{.040}   & \textbf{.904}\\
    \bottomrule
  \end{tabular*}
  
  \label{tab:sota}
\end{table*}

\subsection{Text-to-Mask Generator}
The Text-to-Mask Generator serves as the crucial bridge between instance-specific text prompts derived from MSD-CoT and the segmentation masks, enabling the efficient transformation of semantic textual information into accurate visual segmentation outputs. The process involves two core steps: first, the generation of instance-specific visual prompts from hierarchical text prompts through the proposed Region-constrained Dual-stream Visual Prompting (RDVP), and second, the utilization of these visual prompts as input to a promptable Vision Foundation Model (VFM), such as SAM \cite{kirillov2023segment}, to generate the segmentation masks. To further enhance segmentation quality, a coarse-to-fine strategy is employed, structuring the mask generation into two sequential stages.

\subsubsection{Region-constrained Dual-stream Visual Prompting}
To effectively translate instance-specific text prompts into discriminative visual guidance, we introduce the RDVP module. The RDVP explicitly constrains the selection of foreground and background visual prompts within object-specific bounding boxes. Previous visual prompting methods \cite{hu2024relax,hu2024leveraging} relying on global consensus heatmaps often introduce semantic discrepancy and spatial separation, particularly problematic in highly camouflaged scenarios, as discussed in Section \ref{sec: introduction}. The RDVP overcomes the \textbf{semantic discrepancy} by separately generating independent foreground and background heatmaps using a VLM, such as Spatial CLIP \cite{hu2024relax}, guided by phrase-level ($P_i^f, P_i^b$) or word-level ($W_i^f, W_i^b$) text prompts:
\begin{align}
    H_i^{f}, H_i^{b} = VLM(X, P_i^{f/b}~\text{or}~W_i^{f/b}).
\end{align}

Subsequently, adaptive point selection is conducted separately within each heatmap, restricting sampled visual points to high-confidence regions strictly inside the bounding box $B_i^{'}$ (or $B_i$) for overcoming \textbf{spatial separation}. Specifically, the foreground and background points ($p_i^f, p_i^b$) are selected via:
\begin{align}
    p_i^f &= \{ (x, y) \mid \mathcal{N}(H_i^{f} \mid B_i^{'}~\text{or}~B_i)[x, y] \geq 0.9 \},\\
    p_i^b &= \{ (x, y) \mid \mathcal{N}(H_i^{b} \mid B_i^{'}~\text{or}~B_i)[x, y] \geq 0.9 \},
\end{align}
where $\mathcal{N}(\cdot)$ represents a normalization function, ensuring confidence values are rescaled within a standard range (\textit{i.e.}, [0,1]). The set notation $\{(x, y) \mid \cdot\}$ explicitly denotes the selection of spatial coordinates corresponding to heatmap values exceeding a threshold of 0.9, thereby filtering out unreliable points.

\subsubsection{Coarse-to-Fine Segmentation via VFM}
The instance-specific visual prompts generated by RDVP are subsequently utilized as guidance inputs to the VFM for producing segmentation masks. Formally, the segmentation masks at the coarse ($M_i^{'}$) or fine-grained stage ($M_i$) are generated as: 
\begin{align} 
    M_i^{'}~\text{or}~M_i = VFM(X, p_i^f, p_i^b, B_i^{'}~\text{or}~B_i),
\end{align}
where $B_i^{'}$ is the coarse bounding box predicted by MSD-CoT, providing a preliminary spatial constraint around the camouflaged object. The refined bounding box $B_i$ is derived from the coarse segmentation mask $M_i^{'}$ via the MaxIOUBox operation \cite{hu2024relax}, which selects the box with the highest IoU value with the mask, ensuring tighter spatial alignment for subsequent fine-grained refinement. To achieve robust and precise segmentation, we employ a hierarchical coarse-to-fine pipeline structured into two sequential stages, as shown in Figure \ref{fig: framework}. In the \textbf{coarse stage}, phrase-level text prompts generate initial instance-specific visual prompts via RDVP, resulting in a preliminary segmentation mask $M_i^{'}$ and a refined bounding box $B_i$. Subsequently, in the \textbf{fine-grained stage}, the refined bounding box $B_i$ and the more discriminative word-level text prompts are utilized to provide enhanced instance-specific visual prompts for the VFM. This fine-grained refinement step progressively refines the segmentation mask $M_i$ to achieve pixel-level accuracy.

\begin{figure*}[t]
\centering
\includegraphics[width=1.0\textwidth]{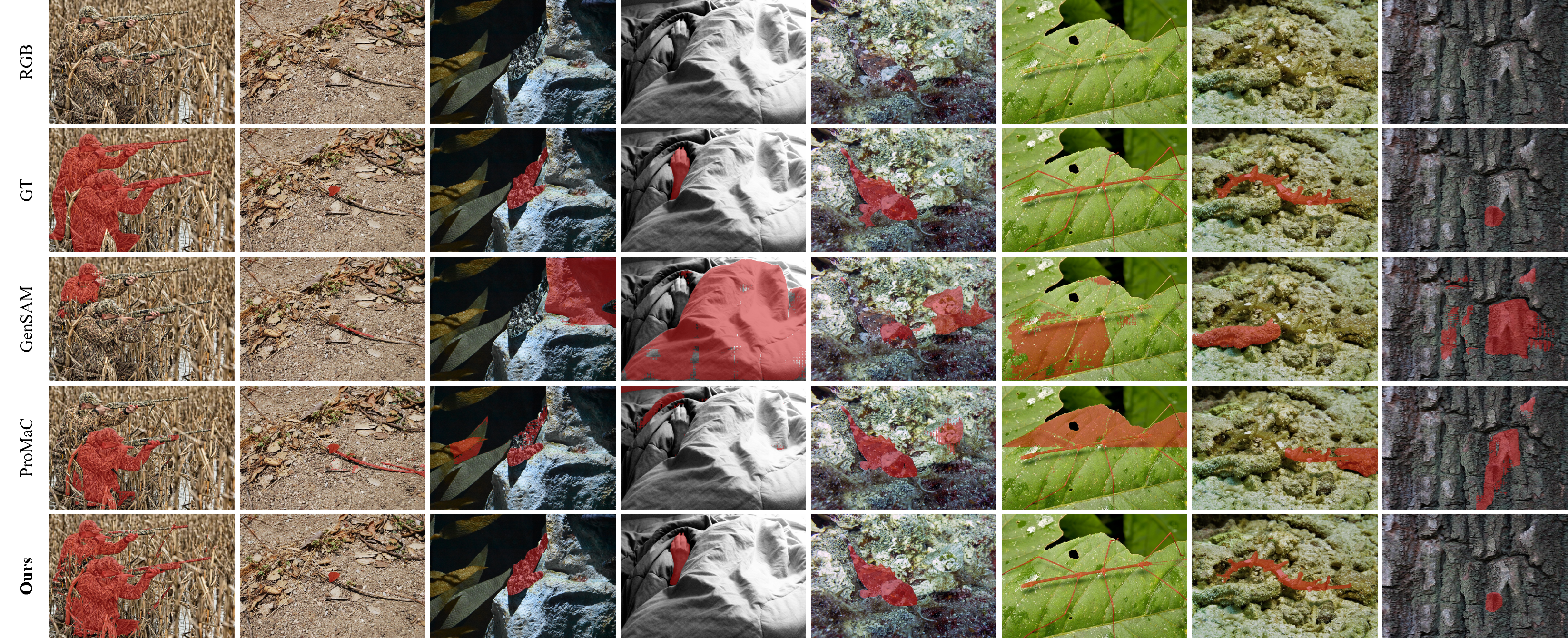} % Reduce the figure size so that it is slightly narrower than the column.
\caption{
Qualitative comparison of the proposed RDVP-MSD with two main task-generic promptable segmentation methods.
}
\label{fig: qualitativeComparison}
\end{figure*}

\subsection{Self-Consistency Mask Selection}
\label{sec: maskSelection}

Existing methods \cite{hu2024relax,hu2024leveraging} often rely on iterative, cycle-generation strategies to produce segmentation masks, repeatedly integrating previous iteration outputs (\textit{e.g.}, heatmaps \cite{hu2024relax} or masks \cite{hu2024leveraging}) back into the original image. However, this iterative dependency typically incurs substantial computational overhead. Inspired by the concept of self-consistency \cite{wang2023selfconsistency,chen2024self} employed in large language models — where the consistency among multiple answers to identical or semantically similar questions is evaluated to enhance reliability — we propose the Self-Consistency Mask Selection mechanism to obtain robust segmentation predictions without iterative dependency efficiently.

Specifically, recognizing that the instance-specific text prompt generation described in Section \ref{sec: MSD-CoT} is inherently stochastic due to its probabilistic nature, we exploit this randomness by performing multiple independent repetitions (denoted as $I$, set to a default value of 3). Each repetition independently generates instance-specific text prompts using semantically equivalent but diverse task-generic synonyms for \textit{camouflaged object}, thus producing multiple candidate segmentation masks $M_i$. These repetitions are mutually independent and thus amenable to parallel execution, significantly enhancing inference efficiency compared to sequential iteration-based approaches.

After obtaining multiple candidate masks, we select the most representative segmentation mask by evaluating their mutual consistency. Formally, the final selected mask index $i^{\star}$ is determined by minimizing the difference between each individual mask $M_i$ and the mean mask computed across all repetitions: 
\begin{align} 
i^{\star} = \arg\min_{i}\left( \left| M_i - \frac{\sum_{i}(M_1,\dots,M_I)}{I}\right| \right), 
\end{align}
where $M_i$ denotes the mask generated by the $i$-th repetition. This consensus-based selection criterion ensures the final segmentation result $M_{i^{\star}}$ exhibits enhanced robustness and accuracy, effectively mitigating uncertainty induced by stochastic text prompt generation.

\section{Experiments}

\subsection{Experiment Settings}
\subsubsection*{Datasets}
To comprehensively assess the performance of the RDVP-MSD model, we conducted experiments using three widely recognized datasets for Camouflaged Object Segmentation (COS): COD10K \cite{fan2020camouflaged}, CAMO \cite{le2019anabranch}, and CHAMELEON \cite{skurowski2018animal}. COD10K is currently the largest COS dataset, containing $5,066$ web-sourced images categorized into $10$ super-classes and $78$ sub-classes. The CAMO dataset comprises $1,250$ images of camouflaged objects, divided into eight categories. The CHAMELEON dataset includes $76$ images for evaluation purposes. Following the evaluation protocols used in previous studies \cite{hu2024relax, hu2024leveraging, tang2024chain}, we tested our model using $2,026$ images from COD10K, $250$ images from CAMO, and $76$ images from CHAMELEON.

\subsubsection*{Implementation details}
For the MLLMs, we use LLaVA-1.5-13B \cite{liu2024improved} for the experiments. For the VLMs, we choose the CLIP \cite{radford2021learning} of the CS-ViT-L/14@336px version. For the VFMs, we deploy the HQ-SAM \cite{ke2023segment} based on the ViT-H version. Our method operates in an entirely train-free test-time adaptation mode. The default value for $I$ is $3$, which means our method repeats the tests $3$ times in parallel. This implies that using $3$ times the computational resources will reduce the single-image inference time reported in Figure \ref{fig: qualitative} to approximately one-third of the original time without compromising performance. All experiments were conducted on two NVIDIA GeForce RTX 3090 GPUs with 24 GB of memory, except for the single-image inference time shown in Figure \ref{fig: qualitative}, where ProMaC \cite{hu2024leveraging} requires at least $3$ RTX 3090 GPUs for reproduction.

\subsubsection*{Evaluation metrics}

In line with prior work \cite{hu2024relax, hu2024leveraging, tang2024chain}, we use four widely recognized metrics to evaluate the performance of our model These include the Structure-measure ($S_{\alpha}$) \cite{fan2017structure}, the adaptive F-measure ($F_{\beta}$) \cite{achanta2009frequency}, the mean absolute error ($M$), and the mean E-measure ($E^{\phi}_m$) \cite{fan2018enhanced}. High-performing COS methods generally achieve higher $S{\alpha}$, $F_{\beta}$, and $E^{\phi}_{m}$ values, along with a lower $M$ score.

\subsection{Comparison with State-of-the-Art Methods}
\subsubsection*{Quantitative Comparison}
We benchmark RDVP-MSD against state-of-the-art methods across three COS datasets under different supervision paradigms, as shown in Table \ref{tab:sota}. In the task-generic prompt setting, our approach achieves the highest $S_{\alpha}$, $F_{\beta}$, $M$, and $E^{\phi}_m$ across all datasets, outperforming prior methods such as ProMaC and GenSAM by substantial margins. Specifically, RDVP-MSD surpasses the second-best method by +6.8\% in $S{\alpha}$, +7.2\% in $F_{\beta}$, and +19.0\% in $M$ on CAMO, while maintaining superior performance across COD10K and CHAMELEON. Compared with point-based or scribble-based weakly supervised methods, RDVP-MSD achieves competitive performance without any form of supervision or training, underscoring its test-time adaptation capabilities and demonstrating significant performance gains in the absence of labeled data. This demonstrates that our method is highly accurate and capable of adapting effectively in real-world scenarios where labeled data is scarce.

\subsubsection*{Qualitative Comparison}
Figure \ref{fig: qualitativeComparison} qualitatively compares RDVP-MSD with leading task-generic promptable segmentation methods. Our approach consistently produces more precise object boundaries, effectively distinguishing camouflaged objects from complex backgrounds while reducing segmentation noise. Compared to GenSAM and ProMaC, RDVP-MSD demonstrates enhanced robustness in highly cluttered or low-contrast scenes, avoiding over-segmentation and under-segmentation artifacts. The results illustrate how our coarse-to-fine prompting strategy refines segmentation masks progressively, capturing fine object details even in the most challenging camouflage scenarios.

\subsection{Ablation Study}
\subsubsection*{Effectiveness of the Modules}
\begin{table} % * 表示这个表格在双栏模板中使用单栏
    \caption{Ablation study on the effectiveness of RDVP-MSD components, demonstrating the performance impact of each proposed module.}
  \centering
  \tabcolsep=0.005\linewidth
    \begin{tabular}{l|cccc|cccc}
    \toprule
         \multirow{2}{*}{Method's Variants} &
         \multicolumn{4}{c|}{COD10K-TEST} &
         \multicolumn{4}{c}{CAMO-TEST}
         \\ 
         \cline{2-9}
         & $S_\alpha$↑ & $F_\beta$↑ & $M$↓ & $E_{m}^\phi$↑
         & $S_\alpha$↑ & $F_\beta$↑ & $M$↓ & $E_{m}^\phi$↑
        \\ 
\toprule
(1) wo MSD-CoT\&RDVP& .795   & .718   & .054   & .854&.756    &.722    &.106    &.818       \\
(2) wo MSD-CoT& .823   & .770   & .042   & .880& .790   & .776   & .089   & .850      \\
(3) wo RDVP&.808    &.738    &.046    & .866&.772    &.748    &.097    &.833       \\

(4) wo $TMG_1$&.814    &.757    &.046    &.872&.772    &.754    &.106    &.833        \\
(5) wo $TMG_2$&.818    &.764    &.045    &.873&.787    &.769    &.091    &.846        \\
\rowcolor{myColor}
(6) Ours& \textbf{.825}   & \textbf{.775}   & \textbf{.038}   & \textbf{.877}&\textbf{.796}   & \textbf{.785}   & \textbf{.081}   & \textbf{.848}      \\
    \bottomrule
  \end{tabular}
  
  \label{tab:ablationForModules}
\end{table}

As shown in Table \ref{tab:ablationForModules}, we perform an ablation study to evaluate the impact of different components on the performance of RDVP-MSD. Setting (1) serves as the baseline, where neither MSD-CoT nor RDVP is used, similar to existing task-generic promptable methods. Setting (2) shows that removing MSD-CoT results in a performance drop, highlighting the importance of phrase disentanglement. Setting (3) demonstrates that replacing RDVP with consensus heatmaps significantly reduces performance, emphasizing the need for independently extracting foreground and background points within the camouflaged object region. Settings (4) and (5) confirm the necessity of the coarse-to-fine segmentation process using phrase-level and word-level prompts. Finally, Setting (6) shows that RDVP-MSD outperforms all other variants. RDVP is the key component contributing to the most significant performance improvement, thus demonstrating its critical role in the model's effectiveness.

\subsubsection*{Effectiveness of RDVP Module}
\begin{table} % * 表示这个表格在双栏模板中使用单栏
    \caption{Ablation experiment of the two main strategies of the RDVP module.}
  \centering
  \tabcolsep=0.012\linewidth
    \begin{tabular}{c|cc|cccc|cccc}
    \toprule
         \multirow{2}{*}{Settings} &
         \multicolumn{2}{c|}{RDVP} &
         \multicolumn{4}{c|}{COD10K-TEST} &
         \multicolumn{4}{c}{CAMO-TEST}
         \\ 
         \cline{2-11}
         &DS & RC
         & $S_\alpha$↑ & $F_\beta$↑ & $M$↓ & $E_{m}^\phi$↑
         & $S_\alpha$↑ & $F_\beta$↑ & $M$↓ & $E_{m}^\phi$↑
        \\ 
\toprule
(1) & & &.808    &.738    &.046    & .866&.772    &.748    &.097    &.833       \\

(2) & \checkmark &     &.820    &.762    &.044    &.875    &.776    &.758    &.097    &.842\\
(3) & &\checkmark  &.821    &.768    &.039    &.877&.784    &.770    &.098    &.844        \\
\rowcolor{myColor}
(4) & \checkmark & \checkmark & \textbf{.825}   & \textbf{.775}   & \textbf{.038}   & \textbf{.877}&\textbf{.796}   & \textbf{.785}   & \textbf{.081}   & \textbf{.848}      \\
    \bottomrule
  \end{tabular}
  
  \label{tab:ablationForRDVP}
\end{table}

As shown in Table \ref{tab:ablationForModules}, the RDVP module is the most critical factor influencing performance, and thus, we use the setting (3) of Table \ref{tab:ablationForModules} as the baseline to evaluate the impact of the two primary strategies in RDVP. The results are presented in Table \ref{tab:ablationForRDVP}. "DS" and "RC" refer to the dual-stream and the region-constrained strategies for generating instance-specific visual prompts, respectively. Settings (1) and (2) demonstrate that extracting foreground and background points from separate foreground/background heatmaps significantly improves performance compared to prior methods that rely on global consensus heatmaps, which sample instance-specific visual prompts from two extreme regions. Settings (1) and (3) highlight that obtaining instance-specific visual prompts within the camouflaged object bounding box, as opposed to using the entire image, more effectively identifies background points that are highly similar to the camouflaged object, thereby enhancing the ability to distinguish it from the background. Finally, setting (4) shows that by combining the strengths of settings (2) and (3), our model achieves the best performance, demonstrating the effectiveness of the region-constrained dual-stream strategy in accurately capturing and distinguishing foreground-background relationships within challenging camouflaged environments.

\subsubsection*{Effectiveness of Repetition Number}
\begin{table} % * 表示这个表格在双栏模板中使用单栏
    \caption{Ablation study on the influence of repetition number $I$ in the proposed Self-Consistency Mask Selection.}
  \centering
  \tabcolsep=0.015\linewidth
    \begin{tabular}{c|cccc|cccc}
    \toprule
         \multirow{2}{*}{Repeat} &
         \multicolumn{4}{c|}{COD10K-TEST} &
         \multicolumn{4}{c}{CAMO-TEST}
         \\ 
         \cline{2-9}
         & $S_\alpha$↑ & $F_\beta$↑ & $M$↓ & $E_{m}^\phi$↑
         & $S_\alpha$↑ & $F_\beta$↑ & $M$↓ & $E_{m}^\phi$↑
        \\ 
\toprule
1&.794    &.727    &.053    &.851&.762    & .746   &.102    &.822        \\
2&.815    &.759    &.042    &.870&.769    &.749    &.096    &.819        \\
\rowcolor{myColor}
3& \textbf{.825}   & \textbf{.775}   & .038   & \textbf{.877}&\textbf{.796}   & \textbf{.785}   & \textbf{.081}   & \textbf{.848}      \\
4&.819    &.769    &.041    & .872&.779    &.762    &.092    &.830       \\
5&.822    &.771    &\textbf{.037}    &.876&.765    &.740    &.092    &.817        \\
6&.822    &.770    &.038    &.871&.770    &.749    &.092    &.822        \\
    \bottomrule
  \end{tabular}
  
  \label{tab: repetitions}
\end{table}

We perform an ablation study to analyze the impact of the hyperparameter $I$, representing the number of parallel repetitions employed in the Self-Consistency Mask Selection module. As shown in Table \ref{tab: repetitions}, increasing the repetition number initially improves segmentation performance due to enhanced mask stability and reduced uncertainty. Optimal performance is achieved at $I=3$, where the model consistently attains the best segmentation accuracy across COD10K and CAMO datasets. However, further increasing repetitions provide negligible accuracy gains, validating our default setting of $I=3$ for practical settings.
\section{Conclusion}
In this work, we introduce RDVP-MSD, a novel training-free test-time adaptation framework that explicitly mitigates the semantic ambiguity arising from instance-specific text prompts generation and mitigates the semantic discrepancy as well as spatial separation encountered during instance-specific visual prompts extraction within task-generic promptable segmentation scenarios for camouflaged objects. Leveraging the proposed MSD-CoT, RDVP-MSD progressively refines instance-specific text prompts, while our RDVP independently obtains the instance-specific visual prompts within spatial constraints. Extensive experiments demonstrate that RDVP-MSD achieves state-of-the-art segmentation accuracy across COS standard benchmarks with substantially improved efficiency without any training or supervision, thus establishing a new paradigm for efficient and precise camouflaged object segmentation.

\begin{acks}
This work is supported in part by the Science and Technology Innovation Plan of Shanghai Science and Technology Commission under grant No.22511106005. We appreciate the High Performance Computing Center of Shanghai University, and Shanghai Engineering Research Center of Intelligent Computing System for providing computing resources and technical support.
\end{acks}

\bibliographystyle{ACM-Reference-Format}
\bibliography{sample-base}

\end{document}